\documentclass[11pt,a4paper]{article}
\usepackage{import}

\usepackage{booktabs}
\usepackage[hyperref]{eacl2021}
\usepackage{times}
\usepackage{latexsym}

\usepackage[T4,T1]{fontenc}
\usepackage{newunicodechar}
\newenvironment{tfour}{\fontencoding{T4}\selectfont}{}
\usepackage{tikz}
\usepackage{tikz-qtree}
\usepackage{tikzsymbols}
\usepackage{array}
\usepackage{wrapfig}
\usepackage{multirow}
\usepackage{tabularx}
\usepackage{booktabs}
\usepackage{graphicx}
\usepackage{subcaption}
\usepackage{enumitem}
\usepackage{amssymb}
\usepackage{amsmath}
\usepackage{comment}
\usepackage{makecell}
\usepackage{multirow}
\usepackage{amsfonts}
\usepackage{booktabs}
\usepackage{lipsum,adjustbox}
\usepackage{xcolor}
\usepackage{pifont}
\newcommand{\cmark}{\ding{51}}%
\newcommand{\xmark}{\ding{55}}%

\makeatletter
\newcommand{\labeltext}[2]{%
  \@bsphack
  \csname phantomsection\endcsname 
  \def\@currentlabel{#1}{\label{#2}}%
  \@esphack
}

\usepackage{xcolor}
\usepackage{etoolbox} 

\makeatletter 
\patchcmd{\maketitle}%
  {\def\@makefnmark{\rlap{\@textsuperscript{\normalfont\@thefnmark}}}}%
  {\def\@makefnmark{\rlap{\@textsuperscript{\normalfont\color{red}\@thefnmark}}}}%
  {}
  {}
\makeatother 

\usepackage{breakurl}
\usepackage{microtype}

\newunicodechar{ɔ}{\fon}
\newunicodechar{ɖ}{\fon}

\aclfinalcopy 

\definecolor{mygray}{RGB}{133,121,89}

\title{OkwuGbé: End-to-End Speech Recognition for Fon and Igbo}

\author{Bonaventure F. P. Dossou\thanks{. These authors contributed equally to this work.} \\
  Jacobs University Bremen \\
  \texttt{f.dossou@jacobs-university.de} \\\And
  Chris C. Emezue\footnotemark[1] \\
  Technical University of Munich \\
  \texttt{chris.emezue@tum.de} \\
  
  }
\date{}
\begin{document}
\usetikzlibrary{calc}
\usetikzlibrary{positioning}
\usetikzlibrary{backgrounds}
\usetikzlibrary{fit}
\usetikzlibrary{3d} 
\usetikzlibrary{shapes.misc, positioning}
\definecolor{spectogram}{rgb}{0.94, 0.87, 0.8}
\definecolor{aliceblue}{rgb}{1.0, 0.97, 0.91}
\definecolor{conv}{rgb}{1.0, 0.94, 0.0}
\definecolor{gru}{rgb}{0.03, 0.27, 0.49}
\definecolor{gray}{rgb}{0.91, 0.41, 0.17}
\definecolor{bilstm}{rgb}{0.06, 0.2, 0.65}
\definecolor{spec}{rgb}{0.94, 0.92, 0.84}
\definecolor{fcl}{rgb}{0.86, 0.86, 0.86}

\maketitle

\begin{abstract}
Language is inherent and compulsory for human communication. Whether expressed in a written or spoken way, it ensures understanding between people of the same and different regions. With the growing awareness and effort to include more low-resourced languages in NLP research, African languages have recently been a major subject of research in machine translation, and other text-based areas of NLP. However, there is still very little comparable research in speech recognition for African languages. Interestingly, some of the unique properties of African languages affecting NLP, like their diacritical and tonal complexities, have a major root in their speech, suggesting that careful speech interpretation could provide more intuition on how to deal with the linguistic complexities of African languages for text-based NLP. OkwuGbé is a step towards building speech recognition systems for African low-resourced languages. Using Fon and Igbo as our case study, we conduct a comprehensive linguistic analysis of each language and describe the creation of end-to-end, deep neural network-based speech recognition models for both languages. We present a state-of-art ASR model for Fon, as well as benchmark ASR model results for Igbo. Our linguistic analyses (for Fon and Igbo) provide valuable insights and guidance into the creation of speech recognition models for other African low-resourced languages, as well as guide future NLP research for Fon and Igbo. The Fon and Igbo models source code have been made publicly available.

\end{abstract}

\section{Introduction}

\begin{center}
$OkwuGb\acute{e} =  \underset{Igbo}{Okwu}(speech) +  \underset{Fon}{Gb\acute{e}}(languages)$
\end{center}
$OkwuGb\acute{e}$, the union of two words from Igbo ($Okwu$) and Fon ($Gb\acute{e}$) means the speech of languages, and signifies studying, and integrating automatic speech recognition to several African languages in an effort to unify them. African languages in the past decade received very little to no research in natural language processing (NLP) \cite{joshi-etal-2020-state,andrew_caines_geographic_2019}, prompting recent efforts geared towards improving the state of African languages in NLP \cite{orife2020masakhane,abbott2018neural,siminyu2020ai4d,nekoto2020participatory}. However, there are few works being done on speech for these African languages, as more emphasis is being placed on their text. Due to the largely acoustic nature of African languages (mostly tonal, diacritical, etc), a careful speech analysis of African languages could provide better insight for text-based NLP involving African languages, as well as supplement the textual data needed for machine translation or language modelling. This is what inspired $OkwuGb\acute{e}$ and the focus on automatic speech recognition.

Automatic speech recognition (ASR, or speech-to-text) is a language technology where spoken words are identified, interpreted and converted to text. ASR is changing the way information is accessed, processed, and used. In recent years, ASR achieved state-of-art performances for most western and Asian languages such as English, French, Chinese, Japanese, etc, due to the availability of large quantity of quality speech resources. African languages, on the other hand are still lacking ASR applications. This is mainly due to the lack or unavailability of speech resources for most African Languages (ALs). In this paper, we introduce ASR systems for two low-resourced languages: Fon and Igbo. We show that using end-to-end deep neural networks (E2E DNN) with Connectionist Temporal Classification (CTC) \cite{ctc}, allows us to achieve promising results without using language models (LMs), which usually require huge amounts of data for training. We also demonstrate that leveraging attention mechanism \cite{bahdanau2016neural} improves the performance of acoustic models.

In section \ref{aboutlang}, we give an overview of Fon and Igbo languages. Then we discuss some related work in section \ref{related} and examine the data and data processing techniques we employed in this research in section \ref{aboutdata}. In section \ref{aboutmodel}, we explore models architectures used for our experiments and show our evaluation in section \ref{results}. 

\section{Overview of Fon and Igbo}
In this section, we give an extensive overview of both languages. Table \ref{langcompare} aims to summarise our analysis for the reader.
\label{aboutlang}
\begin{table*}[h!]
\resizebox{\textwidth}{!}{
    \centering
    \begin{tabular}{| p{5cm} | p{5cm} | p{5cm} |}
    \hline
       \textbf{Characteristics}&\textbf{Fon}&\textbf{Igbo}\\
    \hline
    Spoken where &mostly in Benin. Some part of Nigeria and Togo&mostly in southeastern Nigeria. A little bit in the Equatorial Guinea and Cameroon\\
    \hline
    Speakers \cite{ethnologue} & 2.2 million & 22 million\\
    \hline
    Language family tree &\multicolumn{2}{m{7cm} |}{\begin{tikzpicture}
  [
    grow                    = down,
    sibling distance        = 5em,
    level distance          = 2em,
    edge from parent/.style = {draw, -latex},
    every node/.style       = {font=\footnotesize},
    sloped
  ]
\node {Niger-Congo}
     child { node {Atlantic Congo}
      child { node {Volta-Congo}
       child { node {Kwa}
        child { node {Gbe}
         child { node {Fon}
            edge from parent node [above] {} }
              edge from parent node [above] {} }
                edge from parent node [above] {} }
                child { node {Volta-Niger}
        child { node {Igboid}
         child { node {Igbo}
            edge from parent node [above] {} }
              edge from parent node [above] {} }
                edge from parent node [above] {} }
       }}
    ;
\end{tikzpicture}
}\\
    \hline
    Language structure&Isolating language&Agglutinating language\\
    \hline
    Alphabet structure& 32 letters: 22 consonants, 10 vowels&36 letters: 28 consonants, 8 vowels\\
    \hline
    Special alphabets besides Latin&\begin{tfour}\m{o}\end{tfour}, \begin{tfour}\m{d}\end{tfour}, \begin{tfour}\m{e}\end{tfour}, ã, gb, hw, kp, ny, and xw. & ch, gb, gh, gw, kp, kw, nw, ny, and sh\\
    \hline
    Tonal ?&Yes. 3 tones: high (/), low (\textbackslash) and down step (-)&Yes. 4 tones: high tone ($/$), low (\textbackslash), down step ($-$), and down drift ($-$)\\
    \hline
    Phoneme structure&10 vowel phonemes and 22 consonant phonemes. Nasalization is present&28 consonant phonemes and 8 vowel phonemes. Nasalization is present\\
    \hline
    Number of dialects&about 53&about 30\\
    \hline
    Reduplication ? &Yes, especially in deriving nouns, adjectives, and adverbs from verbs. &Yes, sometimes in compounding word formation: e.g., ugbo (vehicle) + igwe (iron) = ugboigwe (locomotive). \\
    \hline
    Code-switching? &No&Yes\\
    \hline
    \end{tabular}
  }
    \caption{Summary analysis of Fon and Igbo}
    \label{langcompare}
\end{table*}
\subsection{Fon}
\label{fon}
Fon (also known as Fongbe) is a native language of Benin Republic, spoken in average by more than 2.2 million people in Benin, in Nigeria, and Togo \cite{ethnologue}. Fon belongs to the \textit{Niger-Congo-Gbe} languages family, and is a tonal, isolating and left-behind language according to \cite{joshi}, with a basic \textit{Subject-Verb-Object} (SVO) word order. There are currently about 53 different dialects of the Fon language spoken throughout Benin \cite{grammaire, fon_phonology, ethnologue}.

Its alphabet is based on the Latin alphabet, with the addition of the letters: \begin{tfour}\m{o}\end{tfour}, \begin{tfour}\m{d}\end{tfour}, \begin{tfour}\m{e}\end{tfour}, and the digraphs gb, hw, kp, ny, and xw. There are 10 vowel phonemes in Fon: 6 said to be $closed$ [i, u, ĩ, ũ], and 4 said to be $opened$ [\begin{tfour}\m{e}\end{tfour}, \begin{tfour}\m{o}\end{tfour}, a, ã]. There are 22 consonants (m, b, n, \begin{tfour}\m{d}\end{tfour}, p, t, d, c, j, k, g, kp, gb, f, v, s, z, x, h, xw, hw, w). Fon has two phonemic tones: $high$ and $low$. $High$ is realized as rising \textit{(low–high)} after a consonant. Basic disyllabic words have all four possibilities: \textit{high-high}, \textit{high-low}, \textit{low-high}, and \textit{low-low}. In longer phonological words, like verb and noun phrases, a $high$ tone tends to persist until the final syllable. If that syllable has a phonemic $low$ tone, it becomes falling \textit{(high–low)}. $Low$ tones disappear between $high$ tones, but their effect remains as a $downstep$. Rising tones \textit{(low–high)} simplify to $high$ after $high$ (without triggering $downstep$) and to $low$ before $high$ \cite{grammaire, fon_phonology}.

Fon makes extensive use of a rich system of tense or aspect markers, express many semantic features by lexical items, and the periphrastic constructions often used are of a more agglutinative nature \cite{gbe_fon}. Fon nominals are generally preceded by a prefix consisting of a vowel (eg. the word a\begin{tfour}\m{d}\end{tfour}ú: 'tooth'). The quality of this vowel is restricted to the subset of non-nasal vowels \cite{fon_phonology, biblio_gbe}.

Reduplication is a morphological process in which the root or stem of a word, or part of it, is repeated. Fon, like the other Gbe languages, makes extensive use of reduplication in the formation of new words, especially in deriving nouns, adjectives, and adverbs from verbs. For instance, the verb \textit{lã}, which means \textit{to cut} (both in Fon and Ewe), is nominalized by reduplication, yielding \textit{lãlã : the act of cutting}. Triplication is used to intensify the meaning of adjectives and adverbs \cite{fon_phonology, biblio_gbe}.

\subsection{Igbo}
Igbo is a native language of the Igbo people, an ethnic group majorly located in the southeastern part of Nigeria, like Abia, Anambra, Ebonyi, Enugu, and Imo states, as well as in the northeast of the Delta state and in the southeast of the Rivers state. Outside Nigeria, it is spoken a little bit in Cameroon and Equatorial Guinea. Igbo belongs to the Benue-Congo group of the Niger-Congo language family and is spoken by over 27 million people \cite{ethnologue}. There are approximately 30 Igbo dialects, some of which are not mutually intelligible. 
To illustrate the complexity of Igbo, we quote \cite{nwaozuzu2008dialects}: "\textit{...almost every community living as few as three kilometers apart has its few linguistic peculiarities. If these tiny peculiarities are isolated and considered to be able to assign linguistic dependence to each of these communities, we shall therefore be boasting of not less than one thousand languages in what we now know as the Igbo language.}"

This large number of dialects and peculiarities inspired the development of a standardized spoken and written Igbo in 1962, called the Standard Igbo \cite{onwu} (which we will refer to when we say "Igbo"). However, studies have shown that there are many sounds (mainly consonants) found in some other dialects of Igbo which are lacking in the Standard Igbo orthography. For example, \citet{achebe2011composite} discovered about 50 unique speech sounds in Igbo.
Morphologically, Igbo is an agglutinating language, with a compounding word formation: e.g., ugbo (vehicle) + igwe (iron) = ugboigwe (locomotive). Igbo also uses reduplication like Fon. Igbo has 28 consonants and 8 vowels, totalling 36 letters of the alphabet. 

The sound system of Igbo consists of eight vowel phonemes, and 28 consonant phonemes \cite{ikekeonwu1999igbo}. There are four different types of tones in Igbo language \cite{igbochin}. They include: High tone ($/$), Low tone (\textbackslash), Down step ($-$) \cite{10.2307/416372}, Down drift ($-$). Down drift is only observed in Igbo sentences because one can raise or lower the pitch before a sentence is completed. Tone is an integral part of a word in Igbo. It is the interface of phonology and syntax in Igbo because it performs both lexical and grammatical functions \cite{igbophonemes}. Igbo has three syllable types: consonant + vowel (the most common syllable type), vowel or syllabic nasal.\raggedbottom

Code-switching, the act of “alternation of two languages during speech” \cite{Poplack1979SometimesIS}, is very common among Igbo-English bilingual speakers, making it an interesting feature for speech recognition research. Therefore, we will go deeper into it.

\citet{obiamalu,Obiamalu2010MotivationsFC} did an extensive research on code-switching among Igbo speakers, where they classified it into three types: borrowing, quasi-borrowing and true code-switching (see Table \ref{codesw}). Borrowing in Igbo arises when words from English are inserted into Igbo during speech and the words go through phonological and morphological transformation (mark -> \textit{maakigo}, table -> \textit{tebulu}). This is usually because the speaker can not quickly find the Igbo equivalent of the word or such equivalent does not exist. This is illustrated by 1 and 2. In quasi-borrowing, the Igbo equivalents of the English words exist, but the English words are more often used by both monolinguals and bilinguals. It may or may not be assimilated into Igbo, like in borrowing. This is illustrated by 3 and 4. The third situation, called true code-switching, occurs when the speaker purposely chooses to use the English word, even though the Igbo equivalent is known and always used. This is most common among Igbo-English bilinguals. 5 and 6 are good examples.

\begin{table}[h!]
\resizebox{\columnwidth}{!}{
    \centering
    \begin{tabular}{| p{2cm} | p{5cm} | p{6cm} |}
    \hline
       \thead{\textbf{Type}}&\thead{\textbf{Examples (Igbo | \color{mygray}{English})}}&\thead{\textbf{Explanation}}\\
    \hline
    \multirow{2}{2cm}{borrowing}&$1.$\d{O} maakigo (mark) ule ah\d{u}. | \color{mygray}{He has marked the examination}  & \multirow{2}{6cm}{The words ‘mark’ and ‘table’ had been borrowed and assimilated into Igbo because there are not readily available in Igbo.}\\
                &$2.$\d{O} d\d{i} na tebulu (table) | \color{mygray}{It is on the table } & \\\hline

    \multirow{2}{2cm}{quasi-borrowing}&$3.$Obi z\d{u}r\d{u} car \d{o}h\d{u}r\d{u}. | \color{mygray}{Obi bought a new car} & \multirow{2}{6cm}{Even though Igbo has words for ‘car, some bilinguals still use English words.}\\
                &$4.$\d{O}bi z\d{u}r\d{u}  \d{u}gb\d{o}ala \d{o}h\d{u}r\d{u}. | \color{mygray}{Obi bought a new car} & \\\hline
               
    \multirow{2}{2cm}{true code-switching}&$5.$Fela \textit{na e}criticize \textit{onye \d{o}b\d{u}la}.| \color{mygray}{Fela criticizes everybody} & \multirow{2}{6cm}{These cases are true code-switching because the Igbo words for ‘criticize’, ‘turn’, ‘water’ and ‘wine’ are readily available in Igbo, but the speaker chooses to use the English equivalents.}\\
                &$6.$Jesus turn\textit{\d{u}r\d{u}} water \textit{\d{o} gh\d{o}r\d{o}} wine. | \color{mygray}{Jesus turned water into wine} & \\\hline
    \end{tabular}
  }
    \caption{Code-switching types and examples. Adapted from \cite{obiamalu}}
    \label{codesw}
\end{table}

\begin{table*}[t!]
\resizebox{\textwidth}{!}{
    
    \begin{tabular}{m{3cm}m{6cm}m{4cm}m{4cm}m{4cm}}
\toprule
    \centering
    Setting &  &Rich-Resource &Low-Resource &Extremely Low-Resource\\
\midrule
\centering
     &pronunciation lexicon &\cmark &\cmark&\xmark\\
     &paired data (single-speaker, high-quality)&dozens of minutes&several minutes&\xmark\\
     &paired data (multi-speaker, high-quality)&hundreds of hours&dozens of hours&several hours\\
    Data & unpaired speech (single-speaker, high-quality)&\cmark&dozens of hours&\xmark\\
    &unpaired speech (multi-speaker,low-quality)&\cmark&\cmark&dozens of hours\\
    &unpaired text & \cmark&\cmark&very few\\

\hline
Related Work &ASR&[\cite{chorowski2014endtoend,chiu2018stateoftheart,44926,deepspeech,Li_Liu_Liu_Zhao_Liu_2019,end_dnn_agglutinative,badhanau,8683307,rosenberg2019speech,Schneider2019}]&[\cite{tjandra2017listening}]& [\textbf{Our Work}, \citet{laleye, xu2020lrspeech,baevski2020effectiveness,liu2020unsupervised,pmlr-v97-ren19a}]\\
\bottomrule
    \end{tabular}
    }
    \caption{Data sources to build ASR models and the corresponding related works in the different settings. Adapted from \cite{xu2020lrspeech}}
    \label{asr_fon_res}
\end{table*}
\section{Related Works}
\label{related}
In this section, we review some related works according to the data resources, the model architectures and the state of ASR research for Fon and Igbo.
\paragraph{Previous works according to data resources:}

\citet{xu2020lrspeech} classified previous works on ASR, according to data resources, into rich-resource, low-resource and unsupervised settings, as shown in Table \ref{asr_fon_res}.

In the rich-resource setting, a large amount of paired speech and text data is available for training. This amounts up to hundreds of hours by multiple speakers. Furthermore, pronunciation
lexicon is also leveraged while training for better results. These are the languages with ASR models already deployed in the industry. English is a main example of this setting. In the low-resource setting, there are only about dozen minutes of single-speaker high-quality paired data, and few hours of multi-speaker low-quality paired data. Compared to the rich-resource setting, these data resources contained fewer paired data.

In the extremely low-resource setting, which is where our work lies, there are little to no paired speech data resources, very low online presence, and sometimes no developed pronunciation lexicon rule or language model to improve ASR models. Some of these languages also contain few unpaired multi-speaker data. This is the case of many African languages.
\paragraph{Previous works according to model architecture:}
While traditional phonetic-based approaches (Hidden Markov Models) have produced considerable results in the past, we focus on end-to-end speech recognition with deep learning \cite{chorowski2014endtoend,badhanau,deepspeech,deepspeech2,44926,chiu2018stateoftheart} because they have been shown to produce better results, with little dependence on hand-crafted features and phoneme dictionaries. 

\citet{badhanau} introduced an end-to-end continuous ASR using a bidirectional recurrent neural network (RNN) encoder with an RNN decoder that aligns the input and the output sequences using the attention mechanism. The model achieved a word error rate (WER) of 18.57\% on the TIMIT data set. \citet{deepspeech,deepspeech2} presented a state-of-the-art ASR system using E2E DNNs. They introduced a system that does not use any hand-designed language component, nor even the concept of "phoneme". Their result was achieved, as the authors stated in their original paper, through a well-optimized RNN training system that uses multiple GPUs, as well as a set of novel data synthesis techniques and language models.

Following the promising features that E2E DNNs offer, \citet{end_dnn_agglutinative} showed in their recent studies that, using them with CTC works without the need for direct inclusion of language models.

\paragraph{State of ASR resources for African languages:}
Open-sourced data for ASR is one of the driving forces of research in any deep learning field, including ASR, because it fosters experimenting, training and developing better models. The Open Speech and Language Resources (OpenSLR)\footnote{\url{http://openslr.org/index.html}} is a platform with open-sourced speech and language resources, such as training corpora for speech recognition for public use. It currently contains many languages (both high and low resources). However, we discovered that of the 2000 African languages, only Yoruba \cite{gutkin-et-al-yoruba2020}, Nigerian Pidjin and four South African languages (Afrikaans, Sesotho, Setswana, isiXhosa) \cite{van-niekerk-etal-2017} are present, as of the period of writing this paper. Furthermore, they contain very few samples (tens, few hundreds of audio hours), compared to their high-resource counterparts (thousands, millions of audio hours). This scarcity of open resources for the development of ASR for low-resourced African languages is one of the major factors affecting the low state of ASR research in African languages. Although there may be some non-open resources for some of these African languages, they come with huge licensing fees, among other limiting factors.

\paragraph{State of ASR research for Fon and Igbo:}
Fon, unlike Igbo, has little to no digital presence. With very few speakers, and almost no online presence, there have been understandably very few tonal analysis or ASR research for this language. The few that exist are mostly by researchers who are native speakers of the language.

To the best of our knowledge, there has only been notable efforts from \citet{laleyetel-01628455,laleye} to build an ASR for Fon, with a word error rate \textit{(WER)} of 14.83\%. This result had been achieved, building two LMs and also only after normalizing and removing the diacritics; whose crucial importance for both performant ASR and neural machine translation (NMT) has been proved by \citet{orife, ffr}. This will be discussed later in section \ref{text_preprocess}. The best model with diacritics scored a \textit{(WER)} of 44.04\%. 

Igbo, on the other hand, has had a lot of tonal and speech analysis research in the past decade, but no public research on E2E DNN ASR, to the best of our knowledge. We opine that this is largely because 1) many old works in the past on Igbo focused solely on tonal analysis \cite{igbochin,igbophonemes}, and 2) there is a lack of open-source speech data to encourage further research on exploring ASR with deep learning methods, which are known to be data-hungry.\raggedbottom
\section{Speech-to-Text Corpora and Data Preprocessing}
\label{aboutdata}
\subsection{Fon Speech-to-Text Corpus}
We got our speech dataset for Fon from the existing Fon speech corpus\footnote{\url{https://paperswithcode.com/dataset/fongbe-speech-recognition}} which was built upon the tedious task of
recording the texts pronounced by native speakers (including
8 women and 20 men) of Fongbe in a noiseless environment. The recordings are sampled at a frequency of \textit{16Khz}. The 28 native speakers have spoken around 1500 phrases (daily conversations domain). These recordings were made with the LigAikuma\footnote{\url{https://lig-aikuma.imag.fr/}} android application. The minimum length of a speech sample is \textit{2 seconds} and the maximum is \textit{5 seconds}, giving us an average of \textit{4 seconds} content length. Overall, there are around \textit{10 hours} of speech data that have been collected.

The global data set has been split into training, validation and test data sets. The training set contains \textit{8 hours} of speech \textit{(8235 speech samples)}, the validation data set contains \textit{1500 speech samples} and finally the test data set contains \textit{669 speech samples}.
The text corpus made of the \textit{1500} sentences used to build the speech data set has been scraped from BéninLangues \footnote{\url{https://beninlangues.com/fongbe}}.
\subsection{Igbo Speech-to-Text Corpus}
\label{igbo_process}
It was very hard to find the data set of Igbo audio samples and their transcripts. We realized that there is a great lacuna: even though there's been much research on Igbo phonology, there has really not been any (public) efforts to gather any speech-to-text data set for it. 

The data set for our experiments on Igbo was got through a license from the Linguistic Data Consortium (LDC2019S16: IARPA Babel Igbo Language Pack) \cite{ldcigbo}. It contains approximately 207 hours of Igbo conversational and scripted telephone speech collected in 2014 and 2015 along with corresponding transcripts. The data set (hereafter called IgboDataset) is made up of telephone calls representing the Owerri, Onitsha, and Ngwa dialects spoken in Nigeria, sampled at 8kHz, with a few sampled at 48kHz. The gender distribution among speakers is approximately equal; speakers' ages range from 16 years to 67 years. The telephone calls were made using different telephones (e.g., mobile, landline) from a variety of environments including the street, a home or office, a public place, and inside a vehicle. The diacritics were originally removed from the transcripts.

Unlike the Fon data set, which was modern and contained very clean audio, IgboDataset had old speech patterns, and contained many noisy audio samples. Therefore, we had to implement a number of cleaning strategies (like filtering based on length of words, upsampling, exploring different mel-spectrograms units and number of Fast Fourier Transform (FFT) bins. The FFT is an algorithm that computes the discrete Fourier transform (DFT, described in section \ref{speech_process}) of a sequence. Our cleaning strategies gave us a reduced data set of $2.5$ hours, which we split into train, dev and test sizes of 4000, 100 and 100 audio samples respectively. To test the importance of our pre-cleaning, we trained the model on both the large uncleaned data set and our cleaned version (results are discussed in section \ref{results}).
\subsection{Data Preprocessing}
\subsubsection{Speech Preprocessing}
\label{speech_process}
Speech signals are made up of amplitudes and frequencies. Amplitudes simply inform about the loudness of the sound recording, nothing informative. To get more information from our speech samples, we decided to map them into the frequency domain. Two of the known techniques, enabling the conversion of speech data from its time domain to its frequency domain, are the Fourier Transformation (FT) and Discrete Fourier Transformation (DFT) \cite{fourier_application,fourier}. 

FT is a mathematical concept that converts a continuous signal from the time domain to the frequency domain. FT decomposes a continuous signal into its frequency components, giving the frequencies present in the signal, and their respective magnitudes. DFT, similary to FT, converts a sequence, considered as a discrete signal, into its frequency components.



However, applying only the FFT just gives frequency values without time information. To make sure we preserve frequencies, time and amplitudes information about the speech samples, in reasonable and adequate range, we decided then to use mel-spectrograms.

The mel-scale is a perceptual scale of pitches judged by listeners to be equal in distance from one another \cite{mel_1}. It is constructed such that sounds of equal distance from each other on the mel-scale, also «sound» to humans as they are equal in distance from one another. A popular formula to convert $f$ hertz (frequencies measures) into $m$ mels is the O'Shaughnessy formula described in O'Shaughnessy book \cite{mel_2}, defined as \[ m = 2595*log_{10}*(1 + \frac{f}{700}) \hspace{0.25cm} (10).\]

A spectrogram \footnote{\url{http://www.glottopedia.org/index.php/Spectrogram}} is an image, a three dimensional (3D) representation that displays how energy frequencies components of the speech change over time. The abscissa represents time, the ordinate axis represents frequency, and amplitudes are shown by the darkness of a precise frequency at a particular time: low amplitudes are represented with a light-blue color, and very high amplitudes are represented by dark red.

There are two types of spectrograms: broad-band spectrograms and narrow-band spectrograms\footnote{\url{http://www.cas.usf.edu/~frisch/SPA3011_L07.html}}.
\begin{itemize}
    \item Broad-band spectrograms have higher temporal resolutions allowing the detection of changes in frequency over small intervals of time. However, they usually do not help making good frequency distinctions, as the time interval for each spectrum is small.
    \item Narrow-band spectrograms have higher frequency resolution, and larger time interval for every spectrum than broad-band spectrograms: this allows the detection of \textit{very small} differences in frequencies. Moreover, they show individual harmonic structures, which are vibration frequency folds of the speech, as horizontal striations.
\end{itemize}
A mel-spectrogram is hence a spectrogram with the mel-scale. In our study, we decided to use narrow-band mel-spectrograms, as input features for the model.\raggedbottom

We used 512 as length of the FFT window, 512 as the hop-length (number of samples between successive frames) and a hanning windows size is set to the length of FFT window.

For handling the audio data, we used the torchaudio utility from PyTorch \cite{paszke2019pytorch}. We used Spectrogram Augmentation (SpecAugment) \cite{Park_2019} as a form of data augmentation: we cut out random blocks of consecutive time and frequency dimensions. Mel-Spectograms were generated from each speech samples with some fine-tuned hyper-parameters:
\setlist{nolistsep}
\begin{itemize}[noitemsep]
    \item sample\text{\char`_}rate: sample rate of audio signal, set to 16000 hertz (16 kHz) for Fon and 8000 hertz (8 kHz) for Igbo. 
    \item n\text{\char`_}mels: number of mel filterbanks, set to 128 for Fon and 64 for Igbo.
\end{itemize}

\begin{table}[t]
\begin{center}
\resizebox{\columnwidth}{!}{%
    \centering
\begin{tabular}{|p{1cm}||p{3cm}|p{3cm}|}
 \hline
 \textbf{Lang.} & \textbf{Source Text (preprocessed text same as the source)} & \textbf{English Translation}\\
 \hline
 Fon & xov\begin{tfour}\m{e}\end{tfour} sin mì tlala k\begin{tfour}\m{e}\end{tfour}nkl\begin{tfour}\m{e}\end{tfour}n bo ná mì nù\begin{tfour}\m{d}\end{tfour}é
& I'm super hungry, I'm starving. Please give me some food\\
\hline
 Fon &a ná gb\begin{tfour}\m{o}\end{tfour} nù \begin{tfour}\m{d}\end{tfour}ú bó gb\begin{tfour}\m{o}\end{tfour} sin nù bó gb\begin{tfour}\m{o}\end{tfour} xó \begin{tfour}\m{d}\end{tfour}\begin{tfour}\m{o}\end{tfour} káká yi azãn t\begin{tfour}\m{o}\end{tfour}n gbè
& You will not be eating, neither drinking nor speaking for the next three days\\
 \hline
\end{tabular}
    }
\caption{\label{sentences} Fon sentences before and after the preprocessing with their English Translations}
\end{center}
\end{table}
\subsubsection{Text Preprocessing}
\label{text_preprocess}
Scholars like \citet{orife} and \citet{ffr} have shown in their studies that keeping the diacritics reduces lexical disambiguation and provides more morphological information to neural machine translation models. Additionally, diacritics relay the pronunciation tone and the sound generated, leading to an improved understanding of the sentences and their contexts.

\begin{table}[h]
\resizebox{\columnwidth}{!}{
    \centering
    \begin{tabular}{||c||c||}
    \toprule
        \textbf{Fon (a)}& \textbf{Igbo (b)}\\
    \midrule
        
\begin{tikzpicture}
  [
    grow                    = down,
    level distance          = 3em,
    edge from parent/.style = {draw, -latex},
    every node/.style       = {font=\footnotesize},
    sloped
  ]
\node {to (ambiguous and uncertain)}
    child { node {tó (\textit{ear})}}
    child { node {tò (\textit{sea})}}
    child { node {tô (\textit{country})}}
    ;
\end{tikzpicture}
& 
\begin{tikzpicture}
  [
    grow                    = down,
    sibling distance        = 5em,
    level distance          = 3em,
    edge from parent/.style = {draw, -latex},
    every node/.style       = {font=\footnotesize},
    sloped
  ]
\node {akwa (ambiguous and uncertain)}
    child { node {\'akw\`a (\textit{cloth})}}
    child { node {\=akw\'a (\textit{egg})}}
    child { node {\'akw\'a (\textit{to cry})}}
    ;
\end{tikzpicture}
 \\
    \hline
    \bottomrule
    \end{tabular}
    }
    \caption{Examples of tonal inflection with Fon and Igbo}
    \label{fig1}
\end{table}

A diacritic is a glyph added to a letter or basic glyph. They appear above or below a letter, or in some other position such as within the letter or between two letters. Some diacritical marks, such as the acute \textit{(´)} and grave \textit{(`)}, are often called accents. In Fon and Igbo, a good example of the importance can be seen in Table \ref{fig1} (a) where we demonstrate that removing diacritics from a word could lead to ambiguity and result in the confusion of the word. Therefore, we preprocessed the textual data without removing the diacritics.
The results of the text data preprocessing on the Fon data set, are presented in Table \ref{sentences}. For Igbo, unfortunately the IgboDataset was originally stripped of its diacritics. Therefore, we were not able to encode any diacritical information.

\section{Models Architectures and Experiments}
\label{aboutmodel}

\begin {figure*}[!h]
\begin{adjustbox}{width=\textwidth}
\begin{tikzpicture}[
box/.style = {draw, rounded corners=15pt, minimum width=68mm, minimum height=10mm,rotate around={90:(-1,0.5)}},
hbox/.style = {draw, rounded corners=15pt, minimum width=38mm, minimum height=35mm},
nbox/.style = {draw, rounded corners=15pt, minimum width=38mm, minimum height=10mm}]
\tikzstyle{connection}=[ultra thick,every node/.style={sloped,allow upside down},draw=\edgecolor,opacity=0.7]


\node[box,align=center,fill=spec] (n1) at (0.5,0) {Mel-Spectrogram \\ (narrow band)};
\node (a) [box] at (2,0) {};
\foreach \i in {1,...,3}
    \filldraw ($(a.west)+(0.0,0.85*\i)$) [conv] circle (3mm);

\filldraw ($(a.west)+(0.0,0.85*4 - 0.2)$) [conv] circle (0.4mm);
\filldraw ($(a.west)+(0.0,0.85*4)$) [conv] circle (0.4mm);
\filldraw ($(a.west)+(0.0,0.85*4+0.2)$) [conv] circle (0.4mm);
\foreach \i in {5,...,7}
    \filldraw ($(a.west)+(0.0,0.85*\i)$) [conv] circle (3mm);
\node[] (c1) at (1.0,-2.2) {};
\node[] (c2) at (2.0,-2.2) {};
\draw[<->] (c1) --node[anchor=north]{CNN layer} (c2);

\node (btmark1) at (1.5,5.2) {};

\node (r) [box] at (3.5,0) {};
\foreach \i in {1,...,3}
    \filldraw ($(r.west)+(0.0,0.85*\i)$) [gray] circle (3mm);

\filldraw ($(r.west)+(0.0,0.85*4 - 0.2)$) [gray] circle (0.4mm);
\filldraw ($(r.west)+(0.0,0.85*4)$) [gray] circle (0.4mm);
\filldraw ($(r.west)+(0.0,0.85*4+0.2)$) [gray] circle (0.4mm);
\foreach \i in {5,...,7}
    \filldraw ($(r.west)+(0.0,0.85*\i)$) [gray] circle (3mm);
\node (r1) [box] at (5.0,0) {} ; 
\foreach \i in {1,...,3}
    \filldraw ($(r1.west)+(0.0,0.85*\i)$) [gray] circle (3mm);

\filldraw ($(r1.west)+(0.0,0.85*4 - 0.2)$) [gray] circle (0.4mm);
\filldraw ($(r1.west)+(0.0,0.85*4)$) [gray] circle (0.4mm);
\filldraw ($(r1.west)+(0.0,0.85*4+0.2)$) [gray] circle (0.4mm);
\foreach \i in {5,...,7}
    \filldraw ($(r1.west)+(0.0,0.85*\i)$) [gray] circle (3mm);
\node (r2) [box] at (6.5,0) {};
\foreach \i in {1,...,3}
    \filldraw ($(r2.west)+(0.0,0.85*\i)$) [gray] circle (3mm);

\filldraw ($(r2.west)+(0.0,0.85*4 - 0.2)$) [gray] circle (0.4mm);
\filldraw ($(r2.west)+(0.0,0.85*4)$) [gray] circle (0.4mm);
\filldraw ($(r2.west)+(0.0,0.85*4+0.2)$) [gray] circle (0.4mm);
\foreach \i in {5,...,7}
    \filldraw ($(r2.west)+(0.0,0.85*\i)$) [gray] circle (3mm);

\node (r3) [box] at (8.0,0) {};
\foreach \i in {1,...,3}
    \filldraw ($(r3.west)+(0.0,0.85*\i)$) [gray] circle (3mm);

\filldraw ($(r3.west)+(0.0,0.85*4 - 0.2)$) [gray] circle (0.4mm);
\filldraw ($(r3.west)+(0.0,0.85*4)$) [gray] circle (0.4mm);
\filldraw ($(r3.west)+(0.0,0.85*4+0.2)$) [gray] circle (0.4mm);
\foreach \i in {5,...,7}
    \filldraw ($(r3.west)+(0.0,0.85*\i)$) [gray] circle (3mm);

\node (r4) [box] at (9.5,0) {};
\foreach \i in {1,...,3}
    \filldraw ($(r4.west)+(0.0,0.85*\i)$) [gray] circle (3mm);

\filldraw ($(r4.west)+(0.0,0.85*4 - 0.2)$) [gray] circle (0.4mm);
\filldraw ($(r4.west)+(0.0,0.85*4)$) [gray] circle (0.4mm);
\filldraw ($(r4.west)+(0.0,0.85*4+0.2)$) [gray] circle (0.4mm);
\foreach \i in {5,...,7}
    \filldraw ($(r4.west)+(0.0,0.85*\i)$) [gray] circle (3mm);
\node (rmark1) at (3.0,-2.2) {};
\node (rmark2) at (9.0,-2.2) {};

\node[box, align=center,fill=fcl] (fcb) at (11,0) {Fully Connected Layer \\ to flatten output from rCNN block};

\node (b1) [box] at (12.5,0) {};
\foreach \i in {1,...,3}
    \filldraw ($(b1.west)+(0.0,0.85*\i)$) [bilstm] circle (3mm);

\filldraw ($(b1.west)+(0.0,0.85*4 - 0.2)$) [bilstm] circle (0.4mm);
\filldraw ($(b1.west)+(0.0,0.85*4)$) [bilstm] circle (0.4mm);
\filldraw ($(b1.west)+(0.0,0.85*4+0.2)$) [bilstm] circle (0.4mm);
\foreach \i in {5,...,7}
    \filldraw ($(b1.west)+(0.0,0.85*\i)$) [bilstm] circle (3mm);
    
\node (b2) [box] at (14.0,0) {};
\foreach \i in {1,...,3}
    \filldraw ($(b2.west)+(0.0,0.85*\i)$) [bilstm] circle (3mm);

\filldraw ($(b2.west)+(0.0,0.85*4 - 0.2)$) [bilstm] circle (0.4mm);
\filldraw ($(b2.west)+(0.0,0.85*4)$) [bilstm] circle (0.4mm);
\filldraw ($(b2.west)+(0.0,0.85*4+0.2)$) [bilstm] circle (0.4mm);
\foreach \i in {5,...,7}
    \filldraw ($(b2.west)+(0.0,0.85*\i)$) [bilstm] circle (3mm);
    
\node (b3) [box] at (15.5,0) {};
\foreach \i in {1,...,3}
    \filldraw ($(b3.west)+(0.0,0.85*\i)$) [bilstm] circle (3mm);

\filldraw ($(b3.west)+(0.0,0.85*4 - 0.2)$) [bilstm] circle (0.4mm);
\filldraw ($(b3.west)+(0.0,0.85*4)$) [bilstm] circle (0.4mm);
\filldraw ($(b3.west)+(0.0,0.85*4+0.2)$) [bilstm] circle (0.4mm);
\foreach \i in {5,...,7}
    \filldraw ($(b3.west)+(0.0,0.85*\i)$) [bilstm] circle (3mm);


\node (bmark1) at (12.0,-2.2) {};
\node (bmark2) at (15.0,-2.2) {};

\node (g1) [box] at (17.0,0) {};
\foreach \i in {1,...,3}
    \filldraw ($(g1.west)+(0.0,0.85*\i)$) [gru] circle (3mm);

\filldraw ($(g1.west)+(0.0,0.85*4 - 0.2)$) [gru] circle (0.4mm);
\filldraw ($(g1.west)+(0.0,0.85*4)$) [gru] circle (0.4mm);
\filldraw ($(g1.west)+(0.0,0.85*4+0.2)$) [gru] circle (0.4mm);
\foreach \i in {5,...,7}
    \filldraw ($(g1.west)+(0.0,0.85*\i)$) [gru] circle (3mm);

\node (g2) [box] at (18.5,0) {};
\foreach \i in {1,...,3}
    \filldraw ($(g2.west)+(0.0,0.85*\i)$) [gru] circle (3mm);

\filldraw ($(g2.west)+(0.0,0.85*4 - 0.2)$) [gru] circle (0.4mm);
\filldraw ($(g2.west)+(0.0,0.85*4)$) [gru] circle (0.4mm);
\filldraw ($(g2.west)+(0.0,0.85*4+0.2)$) [gru] circle (0.4mm);
\foreach \i in {5,...,7}
    \filldraw ($(g2.west)+(0.0,0.85*\i)$) [gru] circle (3mm);

\node (g3) [box] at (20.0,0) {};
\foreach \i in {1,...,3}
    \filldraw ($(g3.west)+(0.0,0.85*\i)$) [gru] circle (3mm);

\filldraw ($(g3.west)+(0.0,0.85*4 - 0.2)$) [gru] circle (0.4mm);
\filldraw ($(g3.west)+(0.0,0.85*4)$) [gru] circle (0.4mm);
\filldraw ($(g3.west)+(0.0,0.85*4+0.2)$) [gru] circle (0.4mm);
\foreach \i in {5,...,7}
    \filldraw ($(g3.west)+(0.0,0.85*\i)$) [gru] circle (3mm);

\node (gmark1) at (16.5,-2.2) {};
\node (gmark2) at (19.5,-2.2) {};

\node (btmark2) at (20.0,5.2) {};


\node[box,fill=fcl] (fc) at (21.5,0) {Fully Connected Layer};
\node[box,fill=spec] (ctc) at (23.0,0) {CTC Loss};


\draw[<->] (bmark1.west) --  node[below, anchor=north]{BiLSTM block} (bmark2.east);
\draw[<->] (rmark1.west) --  node[below, anchor=north]{rCNN block} (rmark2.east);
\draw[<->] (gmark1.west) --  node[below, anchor=north]{BiGRU block} (gmark2.east);
\draw[|-|] (btmark1.west) --  node[above]{Batch Normalization} (btmark2.east);

\node[nbox] (nothing) at  (3.0,-4.5) {Output from previous layer};
\node[hbox, below of=nothing,yshift=-1.6cm,align=center] (ln) {Layer Norm \\ | \\ GeLU \\ |  \\ Dropout \\ | \\ CNN };
\node[hbox, below of=ln,yshift=-3.2cm,align=center] (ln2) {Layer Norm \\ | \\ GeLU \\ |  \\ Dropout \\ | \\ CNN };
\node[below of=ln2, draw, circle,fill=green!40,yshift=-1.55cm] (sum) {$+$};
\node[nbox,below of=sum,yshift=-0.25cm] (nothing2) {Input for next layer};

\draw[<->] (nothing) -- (ln);
\draw [<->] (ln) -- (ln2);
\draw [<->] (ln2) -- (sum);
\draw [<->] (sum) -- (nothing2);
\draw[->](nothing.west) to [out=180,in=180] (sum.west);

 \begin{scope}[on background layer]
\node (rdisp) [draw, rectangle, rounded corners=15pt, fit= (nothing) (nothing2),color=gray,line width=0.5mm,fill=aliceblue!50] {};
\end{scope}
\draw[->] (r) -- (rdisp);

\node (nt) [nbox] at (15.0,-4.0) {Output from previous layer};
\node (lm) [nbox, align=center,below of=nt,yshift=-0.6cm]  {Layer Norm \\ | \\ GeLU};
\node [nbox, minimum width = 58mm,below of=lm,yshift=-0.4cm, align=center] (enc) {Bidirectional GRU \\ (outputs $x$ and hidden state)};
\node [nbox,below of=enc,yshift=-0.4cm] (ench) {Hidden state};
\node [nbox,below of=ench,yshift=-0.4cm] (all) {Alignment score};
\node [nbox,below of=all,yshift=-0.8cm] (attn) {Attention weights};
\node [nbox,below of=attn,yshift=-0.4cm] (con) {Context Vector};
\node[below of=con, draw, circle,fill=pink!40, yshift=-0.5cm,minimum width=6mm] (ct) {};
\node [nbox,below of=ct,yshift=-0.4cm, align=center, minimum width=48mm] (dec) {Input for next layer};

\node (attnlayer) [draw, dashed, fit= (all) (con), minimum width=45mm,color=blue] {};

\node (btn) [right of=attnlayer, xshift=5.1cm] {};
\node (btn1) [right of=enc, xshift=4.5cm,yshift=-0.7cm] {};
\node (btn2) [right of=enc, xshift=1.83cm,yshift=-0.7cm] {};

 \begin{scope}[on background layer]
\node (attnrect) [fit=(nt) (dec),draw, rounded corners=15pt,minimum width=63mm,color=gru,line width=0.5mm,fill=aliceblue!50] {};
\end{scope}

\draw[<->] (nt) -- (lm);
\draw[<->] (lm) -- (enc);
\draw[<->] (attn) -- (con);
\draw[<->] (con) -- (ct);
\draw[<->] (ct) -- (dec);
\draw[<->] (all) -- node[anchor=east]{softmax} (attn);

\draw[->] (btn) -- node[align=center]{Attention \\ mechanism} (attnlayer);
\draw[->] (btn1) -- node[align=center]{RNN-type \\ connection} (btn2);

\draw[->,>=stealth] (enc.west) to [out=180,in=180] node[above,sloped]{concat. with $x$} (ct.west);

\draw[<->,color=purple] (enc.west) to [out=180,in=180] (ench.west);
\draw[<->,color=purple] (enc.east) to [out=0,in=0] (ench.east);

\draw[<->,color=purple] (ench.west) to [out=180,in=180] (all.west);
\draw[<->,color=purple] (ench.east) to [out=0,in=0] (all.east);

\draw[->] (g1) -- (attnrect) {};

\end{tikzpicture}
\end{adjustbox}
\caption{Architecture of the best model for Fon and Igbo, with an expansion of each component of the rCNN block and BiGRU block.}
\label{model}
\end{figure*}
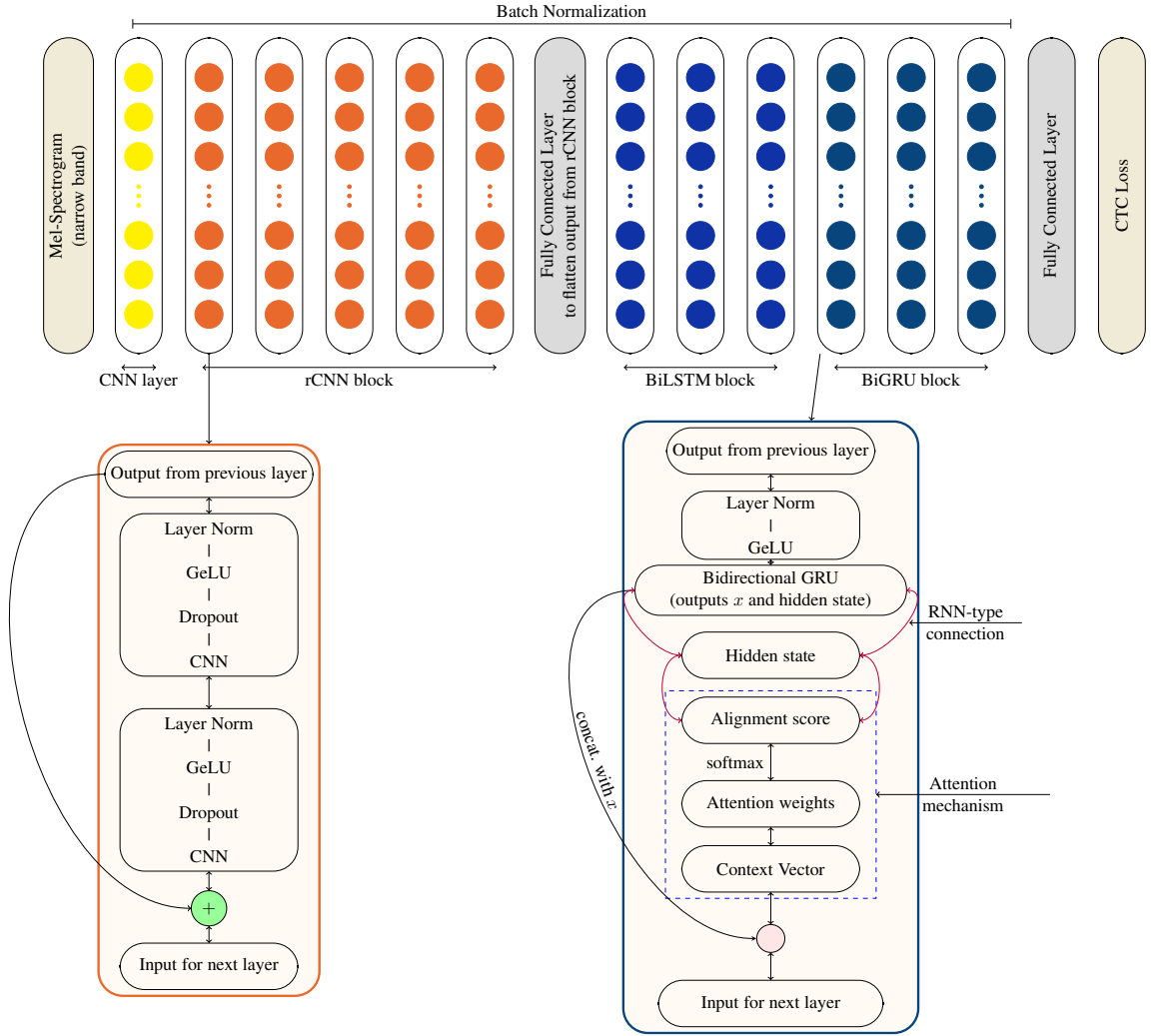


\subsection{Preliminaries}
We take a moment to briefly define the task of ASR mathematically. We have a training set of $n$ samples: $\chi$ = $\left\{ (x^{(1)}, y^{(1)}),(x^{(2)}, y^{(2)}),..., (x^{(n)}, y^{(n)}) \right\}$. Each utterance, $x^{(i)}$, sampled from the training set, is a time-series of length $T^{(i)}$ where every time-slice is a vector of audio features,  $(x_{t}^{(i)}, t = 0,..., T^{(i)}-1$. The goal of ASR is to generate transcripts $\hat{y}^{(i)}$ for each utterance $x^{(i)}$. In order to achieve this, we use an architecture consisting of one or more recurrent neural networks (RNN), since they are best equipped for time-series data. 

In order to generate $\hat{y}^{(i)}$ for a given utterance $x^{(i)}$, the RNN models the probability of picking a character from the character set. To put it mathematically, at each output time-step $t$, the RNN makes a prediction over characters, $p(c_{t}|x^{(i)})$, where $c_{t}$ is either a character of the alphabet (including their diacritics) or the blank symbol. In Fon for example, due to our inclusion of the diacritics, we have $c_{t} = \{ a,b,c,...,z,$ \textit{à, á, ā, ă, è, é, ē, ĕ, ì, í, î, ï, ĭ, ó, ŏ, ò, ū, ŭ, ù, ú, \begin{tfour}\m{o}\end{tfour}, \begin{tfour}\'{\m{o}}\end{tfour}, \begin{tfour}\`{\m{o}}\end{tfour}, \begin{tfour}\u{\m{o}}\end{tfour}, \begin{tfour}\m{d}\end{tfour}, \begin{tfour}\m{e}\end{tfour}, \begin{tfour}\`{\m{e}}\end{tfour}, \begin{tfour}\'{\m{e}}\end{tfour}, \begin{tfour}\u{\m{e}}\end{tfour}, \begin{tfour}\~{\m{e}}\end{tfour}, $fullstop$, $apostrophe$, $comma$, $space$, $blank$}$\}$. Apostrophe, white-space, comma, and full stop characters have been added to denote word boundaries. For Igbo, $c_{t} = \{a,b,...,z,$ \textit{fullstop, apostrophe, comma, blank}$\}$ which is the same as the character set for the English language. This constraint was because the speech dataset for Igbo was already stripped of the diacritics, as explained in section \ref{igbo_process}.

\subsection{Model Architecture}
\label{model_architecture}
Related works have shown that we can increase model capacity, in order to efficiently learn from large speech datasets, by adding more hidden layers rather than making each layer larger. \citet{Graves2013SpeechRW} explored increasing the number of consecutive bidirectional recurrent layers, and \citet{deepspeech2} proposed the Deep Speech2, which among a number of optimization techniques, extensively applied batch normalization \citep{batch_norm} to the deep RNNs.

Furthermore, \citet{chorowski2014endtoend} showed that the use of Badhanau (additive) attention mechanism \cite{bahdanau2016neural} could reduce the phoneme or word error rate (WER) of the ASR model. This is possible because the attention mechanism forces the decoder to make monotonic alignment and hence improve the predictions.

Our model architecture, shown in Figure \ref{model}, draws inspiration from these research findings. While our model at its core is similar to Deep Speech 2, our key improvements are: 
\begin{itemize}
    \item the exploration of the combination of Bidirectional Long Short Term Memory (BiLSTMs) and Bidirectional Gated Recurrent Units (BiGRUs) for low-resource ASR.
    \item the integration of the Badhanau attention mechanism, which effect has been demonstrated on Fon.
\end{itemize}
Our model has two main neural network modules: $N$-blocks of  Residual Convolutional Neural Networks (rCNNs) \cite{he2015deep,resnet} and $M$-blocks each of BiLSTMs and BiGRUs. Each rCNN block is made of two CNN layers, two dropout layers and two normalization layers \cite{Ba2016LayerN} for the CNN inputs. We leveraged the power of convolutional neural networks (CNNs) to extract abstract features by converting speeches into spectrograms. RNNs process the abstract features produced by the rCNNs, step by step, making a prediction for each frame while using context from previous frames. We use BiRNN's \citep{birnns} because we want the context of not only the frame before each timestep, but the following as well. This help the model make better predictions.

In our scenario, BiLSTMs and BiGRUs act respectively as encoder and decoder blocks. Each block produces subsequentially outputs and hidden states fed to the next block. The last hidden state of the last BiGRU block is used to compute the attention weights and the context vector, that will be concatenated to the BiGRU output, to serve as final output. In-between and towards each block output are stacked dropout layers to prevent over-fitting \citep{dropout}.

The output from the models is a probability matrix for characters which will be fed into a greedy decoder. We implemented the greedy decoder suggested by \citet{10.1145/1143844.1143891} to extract what the model believes are the highest probability characters that were spoken. This simple decoder, albeit with no linguistic information, has been shown to produce useful transcriptions \cite{zenkel2017comparison}.

Our model is trained to predict the probability distribution of every character of the alphabet at each timestep from the narrow-band mel-spectrogram we feed it. Traditional ASR models require aligning transcripted text to the speech before the training, and the model is trained to predict specific labels at specific timesteps. However, with the CTC loss function \citep{ctc}, the previously described step is skipped and the model directly learns to align the transcript itself during training. 
\subsubsection{Implementing the attention mechanism}
Let us recall here that we have 5 blocks of rCNNs, 3 blocks each of BiLSTMs and BiGRUs. Bahdanu attention a bit modified and implemented as explained in Figure \ref{model}, in the following steps:
\begin{itemize}
    \item an input $x$ from the stacked blocks of rCNNs is fed to the BiLSTM (encoder), and primarily layer-normalized. The output is then passed through a GeLU activation function, which output is fed to the BiLSTM layers of the current block. Within the current encoder block, each of the BiLSTM layers, produce an output, a hidden state and a cell state. The encoder output is lastly passed to a dropout layer, and is used as input for the next encoder block. The output of the last encoder is used as input for the first BiGRU (decoder) block.
    \item The input of the decoder goes successively through a normalization layer and a GeLU activation function. The output is then fed to BiGRU layers of the current decoder, which produces the decoder output and a hidden state. In common NLP or more specifically in neural machine translation (NMT), the hidden state is a 2-dimensional tensor. However, this is not the case here, since our initial input features from the stack of rCNNs layers are 4-dimensional tensors of shape \textit{(batch, channel, feature, time)}, and the output from the stack of encoders is a 3-dimensional tensor of shape \textit{(batch\text{\char`_}size, number of features, hidden size)}.
    \item Using the hidden state $h$, and the decoder output $x$ shapes, three dense layers are created accordingly using the following PyTorch version pseudo-code:
    \begin{gather*}
    w1 = nn.Linear(x.size[2], x.size[2]//2)\\
    w2 = nn.Linear(h.size[2], h.size[2])\\
    v = nn.Linear(x.size[2]//2, x.size[2])
    \end{gather*}
    Along our manipulation, in the attempt to customize the NMT concept to ours, we encountered few cases were shapes of the decoder output and the hidden state were mismatching. To resolve that, we augmented the hidden state tensor, along the second axis $(axis=1)$ with desirable number of zeros. This was done for two reasons:
    \begin{enumerate}
        \item the concatenation will not affect the original hidden state tensor
        \item $tanh(x|x=0) = 0$: 0 is hence acting like a neutral element.
    \end{enumerate}
    Once we get compatible shapes for both the decoder output and the hidden state tensor, we used them to conventionally compute the attention scores, $s$:
    \begin{gather*}
    m = nn.Tanh()\\
    s = v(m(w1(x)+w2(h))
    \end{gather*}
    \item The attention weights are then computed by passing $s$ through a softmax operation. The context vector, $c_{v}$ , which is got by multiplying the attention weights by $x$, is concatenated to the decoder output $x$ to get the attention features:
    \begin{gather*}
    n = nn.Softmax()\\
    attention\text{\char`_}weights = n(s)\\
    c_{v} = attention\text{\char`_}weights * x\\
    x = concatenate(c_{v}, x)
    \end{gather*}
    \item The output of the current decoder block is the attention output passed through a dropout layer, and taken as input for the next decoder.
\end{itemize}
\subsection{Experiments}
Throughout our experiments, we explored different model architectures with various number of convolutional and bidirectional recurrent layers. The best ASR model, shown on Figure \ref{model} has $5$ blocks of rCNNs, $3$ blocks each of BiLSTMs and BiGRUs, with attention incorporated into each component of the BiGRU block. Also, we used a form of Batch Normalization throughout the model.

We got the best evaluation results with the following hyper-parameters:

\begin{itemize}
   \item max learning\text{\char`_}rate: 5e-4 (for Fon), 3e-4 (for Igbo)
   \item batch\text{\char`_}size: 20 (for Fon), 20 (for Igbo). 
   \item ($N,M$): (5, 3) for Fon and Igbo.
   \item embedding\text{\char`_}size: 512
   \item epochs: 500 (for Fon) and 1000 (for Igbo), with early stopping after 100 epochs.
   \item activation\text{\char`_}function: GeLU \cite{gelu}
   \item optimizer: AdamW \cite{loshchilov2019decoupled} (Fon), Nesterov accelerated descent \cite{Nesterov1983AMF} (Igbo)
\end{itemize}
We used two metrics to evaluate the models: the Character Error Rate (CER) and the WER. WER uses the Levenshtein distance \cite{levenshtein} to compare reference text and hypothesis text in word-level. Even with a low \textit{CER}, the \textit{WER} can be high: hence the lower the \textit{WER}, the better the model. For the training processes, we used the \textit{WER} of the validation data set to select the best weights and parameters.
\section{Results}
\label{results}
\begin{table}[h!]
\resizebox{\columnwidth}{!}{
    \centering
    \begin{tabular}{|c|cc|cc|}
    \toprule
        Models &\multicolumn{2}{c}{Fon}&\multicolumn{2}{|c|}{Igbo (without cleaning)}\\
         (rCNN)& CER & WER & CER &WER\\
    \midrule
         +BiGRUs & 22.0831& 59.66& - & - \\
          +BiLSTMs&24.2783&61.46& - & -\\
          +BiLSTMs+BiGRUs&\textbf{16.9581}&47.05 &56.00 & 64.00\\
           +\textbf{BiLSTMs+BiGRUs+Attn}& 18.7976& \textbf{42.50}& \textbf{50.12} (92.67) &\textbf{55.03} (97.99) \\
           \hline
         \citep{laleye}\footnotemark&-&44.09&-&-\\
    \bottomrule
    \end{tabular}
    }
    \caption{CER (\%), and WER (\%) of different models on Fon and Igbo (original and cleaned) test datasets.}
    \label{asr_fon}
\end{table}
\footnotetext[7]{WER of the best model \textit{with diacritics} from \citep{laleye}}

\begin{table*}[!t]
\begin{center}
\resizebox{0.85\textwidth}{!}{
    \centering
\begin{tabular}{|c|c|}
 \hline
 \bf{Fon Decoded Predictions} & \bf{Fon Decoded Targets}\\
 \hline
 t\begin{tfour}\m{o}\end{tfour} ce xwe y\begin{tfour}\m{o}\end{tfour}y\begin{tfour}\m{o}\end{tfour} din t\begin{tfour}\m{o}\end{tfour}n \begin{tfour}\m{o}\end{tfour} ci gblagadaa & t\begin{tfour}\m{o}\end{tfour} ce xwe y\begin{tfour}\m{o}\end{tfour}y\begin{tfour}\m{o}\end{tfour} din t\begin{tfour}\m{o}\end{tfour}n \begin{tfour}\m{o}\end{tfour} ci gblagadaa\\
\hline
 eo mi sa \textcolor{purple}{aakpan} nu mi & eo mi sa \textcolor{purple}{akpan} nu mi \\
\hline
 fit\begin{tfour}\m{e}\end{tfour} a \textcolor{purple}{gosin} xwe  \textcolor{purple}{yi} gbe & fit\begin{tfour}\m{e}\end{tfour} a \textcolor{purple}{go sin} xwe \textcolor{purple}{yin} gbe\\
\hline
 e kpo kp\begin{tfour}\m{e}\end{tfour}\begin{tfour}\m{d}\end{tfour}\textcolor{purple}{é}&e kpo kp\begin{tfour}\m{e}\end{tfour}\begin{tfour}\m{d}\end{tfour}\textcolor{purple}{e}\\
 \hline
 akw\begin{tfour}\m{e}\end{tfour} \textcolor{purple}{c}\begin{tfour}\m{e}\end{tfour} gbadé j\textcolor{purple}{í} \begin{tfour}\m{d}\end{tfour}axim\begin{tfour}\m{e}\end{tfour} &  akw\begin{tfour}\m{e}\end{tfour} \textcolor{purple}{j}\begin{tfour}\m{e}\end{tfour} gbadé j\textcolor{purple}{i} \begin{tfour}\m{d}\end{tfour}axim\begin{tfour}\m{e}\end{tfour}\\
 \hline
\end{tabular}
    }
\caption{\label{asr_predicitons} Decoded Predictions and Targets of the best Fon ASR Model}
\end{center}
\end{table*}

We present our findings using $5$ blocks of rCNNs with:
\begin{itemize}
    \item $3$ blocks solely of BiGRUs
    \item $3$ blocks solely of BiLSTMs
    \item $3$ blocks each of BiLSTMs and BiGRUs
    \item $3$ blocks each of BiLSTMs and BiGRUs + Attention mechanism.
\end{itemize}    
    
Table \ref{asr_fon} presents the results of different models architectures on the test data set for Fon and Igbo.

\subsection{Results for Fon}
We show that implementing attention mechanism reduced the WER by 5\%. Our Fon ASR model outperformed the current Fon ASR model with diacritics of \citet{laleye}.

Table \ref{asr_predicitons} shows some decoded predictions and targets from the Fon ASR model, which are very similar. Common mistakes (\textcolor{purple}{colored}), happen most often at a character level where a character is either omitted, added or replaced by another one. The native speakers included in this study have testified to the fact that those mismatched words or characters are often practically not distinguishable in speaking. The model source code is open-sourced at: \url{https://github.com/bonaventuredossou/fonasr}
\subsection{Results for Igbo}
An important observation we show in Table \ref{asr_fon} is the effect of the state of audio samples on the model's ability to learn: for our large IgboData with background noise, uneven audio length, low sampling rate, etc, the model found it very difficult to learn the speech representations. Taking time to sieve through the data (in Section \ref{igbo_process}) mitigated this issue by helping the model learn the abstract features better, albeit on a small training set. While the model is currently still training on more epochs (with hope of improving), our preliminary results serve as a benchmark for ASR on Igbo. The source code for the model can be accessed at \url{https://github.com/chrisemezue/IgboASR}.

In Table \ref{asr_fon}, one may observe the large difference between the CER and WER on Fon language, unlike Igbo. We strongly believe that this is due to the fact that the character set for Fon contains all the possible diacritics for each letter of the Fon alphabet, making it extremely large (compared to the set of Igbo characters which had no diacritical information). To further support our claim, a close observation of the targets and predictions in Table \ref{asr_predicitons} reveals that the errors are mostly due to omission or mismatch of diacritics for the characters (`e' predicted instead of `\'{e} ' in row 4 or a space added between `go' and `sin' in row 3 ). 
\section{Future Work}
Our work shows promising results considering the small training sizes, and we have presented a state-of-the-art ASR model for Fon. As future pathway to improve the proposed models, we are exploring approaches like leveraging language models, deeper model structures, transformers and crowd-sourcing/compiling speech-to-text data set for Igbo and Fon.

For Igbo language, the next stage involves incorporating diacritical information in the ASR model. We have begun by gathering new speech dataset which include the diacritics.
\section{Acknowledgement}
\label{acknowledgement}
We are grateful to Professor Graham Neubig of Carnegie Mellon University for coming to our aid by providing us with an Amazon EC2 instance for training our model when we were very low on computational resources. We also thank Dr Frejus Layele for giving us access to the Fon data set, and Dr Iroro Orife, for his guidance on designing the ASR model and cleaning the IgboData.
\bibliography{anthology,eacl2021}
\bibliographystyle{acl_natbib}

\end{document}